\documentclass[runningheads]{llncs}
\usepackage{float}
\usepackage{soul}
\usepackage{threeparttable}
\usepackage{hyperref}
\usepackage{subcaption}
\usepackage{multicol}
\usepackage{multirow}
\usepackage{amsmath}
\usepackage{amssymb}
\usepackage{adjustbox}
\usepackage{textcomp}
\usepackage{graphicx}
\usepackage{xcolor}
\definecolor{blue}{RGB}{0, 128, 255}
\definecolor{red}{RGB}{255,0,0}
\definecolor{green}{RGB}{0,255,0}

\begin{document}
\title{Motion Artifacts Detection in Short-scan Dental CBCT Reconstructions} 

%
\titlerunning{CBCT Motion Detection}
%

\author{Abdul Salam Rasmi Asraf Ali \inst{1,2} \and
Andrea Fusiello\inst{1} \and
Claudio Landi\inst{2} \and
Cristina Sarti\inst{2} \and
Anneke Annassia Putri Siswadi\inst{3,4}
}
\authorrunning{Asraf Ali et al.}
%
\institute{Department of Engineering and Architecture, University of Udine, Italy \and
See Through s.r.l., Brusaporto, Italy \and
ImViA/IFTIM, University of Burgundy, Dijon, France \and
Department of Information Technology, Gunadarma University, Depok, Indonesia
}

\maketitle              
\begin{abstract}
Cone Beam Computed Tomography (CBCT) is widely used in dentistry for diagnostics and treatment planning. CBCT Imaging has a long acquisition time and consequently, the patient is likely to move. This motion causes significant artifacts in the reconstructed data which may lead to misdiagnosis. Existing motion correction algorithms only address this issue partially, struggling with inconsistencies due to truncation, accuracy, and execution speed.  On the other hand, a short-scan reconstruction using a subset of motion-free projections with appropriate weighting methods can have a sufficient clinical image quality for most diagnostic purposes. Therefore, a framework is used in this study to extract the motion-free part of the scanned projections with which a clean short-scan volume can be reconstructed without using correction algorithms. Motion artifacts are detected using deep learning with a slice-based prediction scheme followed by volume averaging to get the final result. A realistic motion simulation strategy and data augmentation has been implemented to address data scarcity. The framework has been validated by testing it with real motion-affected data while the model was trained only with simulated motion data. This shows the feasibility to apply the proposed framework to a broad variety of motion cases for further research.

\end{abstract}

\section{Introduction}
Cone Beam Computed Tomography is a three-dimensional imaging method widely used in dentistry that allows the visualization of the maxillofacial region from any viewpoint \cite{patel2015cone}. The usage of CBCT has increased significantly in various fields such as orthodontics, endodontics, oral surgery, periodontics, and restorative dentistry \cite{small2007cone}. Unlike clinical helical computed tomography (CT), CBCT systems have acquisition time ranging from 5.4 to 40 seconds \cite{nemtoi2013cone}, which is often long enough for significant patient motion to occur. Patients move during the scan for various reasons such as fear of the tube/detector movement \cite{yildizer2017effect}, elderly people having Parkinson's disease \cite{donaldson2013dental}, or simply because it is difficult for someone to stay idle for quite a long time, especially children \cite{spin2015factors}. It is estimated that approximately 21–42\% of the in vivo examinations exhibit motion artifacts \cite{nardi2015metal,spin2015factors}. 

Many efforts have been made to prevent patient motion during CBCT acquisition. Patients are generally immobilized using a head strap, chin rest, and a fixed bite block. However, relying on head fixation devices may not prevent all potential motions \cite{hanzelka2013movement}, since as small as $3mm$ of motion displacement can significantly affect the image quality \cite{spin2018ex}. Several works have been done in the past to compensate for patient motion, e.g.,  based on a 2D-3D registration process \cite{ouadah2017correction,wein2011self}, based on Epipolar consistency \cite{aichert2015epipolar}, based on the optimization of data fidelity terms such as local edge entropy \cite{sisniega2017motion,wicklein2013online}, auto-calibration based approaches \cite{maur2019cbct}, and deep learning based approaches \cite{ko2021rigid}. However, these approaches are only addressing the issue partially, struggling with inconsistencies due to truncated data. Another general drawback of these approaches is the compromise between accuracy and execution time.  Moreover, current manufacturers of maxillofacial CBCT scanners (Sirona Dental Systems, KaVo Dental, VATECH, Carestream Dental) do not offer motion compensation features, excluding one provider who offers partial motion correction (Planmeca CALM \textregistered).

Although many works have been published related to the compensation for patient motion, only a few focused on detecting the motion artifacts in CBCT Imaging. Ens et al. \cite{ens2010automatic} proposed an automatic detection of patient motion in the CBCT projection images using several similarity measures like Structural Similarity Index Measure and Mutual Information. The drawback of this work is that the motion is detected by comparing two consecutive projections at a time which is time-consuming. Also, this method might not be robust to changes in the quality of projections which heavily depends on radiation dose. Sun et al. \cite{sun2020motion} presented a motion artifact detection algorithm based on the Convolutional Neural Network (MADA-CNN) implemented with transfer learning and ensemble modeling. While their method has achieved an AUC-ROC value of 0.966 for motion detection in slices from the reconstructed volumes, the effects on full 3D volumes have not been investigated.  Welch et al. \cite{welch2020automatic} and Arrowsmith et al. \cite{arrowsmith2021automated} used a 3D Convolutional Neural Network (CNN) for detecting dental artifacts directly in the reconstructed CBCT volumes which is much faster than detecting such artifacts in projection images but training a 3D CNN is computationally expensive. Furthermore, artifacts caused due to different reasons need different compensation techniques, and grouping all artifacts together can make it difficult to determine the best approach for each individual case.

Moreover, Parker \cite{parker1982optimal} stated that a short-scan reconstruction of approximately $180^{\circ}$ plus the fan angle with appropriate weighting methods is diagnostically viable for most clinical purposes. The focus of this paper is to extract a part of the scanned CBCT projections, free from motion artifacts, to reconstruct a short-scan volume that can be used directly for diagnosis without the need for a correction step. To achieve this, four short-scan volumes are reconstructed from selected projections to cover the entire $360^{\circ}$ view, and a deep learning-based framework is implemented to classify those volumes as positive (with motion artifacts) or negative (without motion artifacts). A motion simulation strategy has also been implemented to imitate the kinds of motion, patients can experience during CBCT scanning.

\section{Materials and Methods}

\subsection{Data Acquisition}
\begin{figure}[htbp]
    \begin{subfigure}{0.32\textwidth}
    \includegraphics[width=\textwidth]{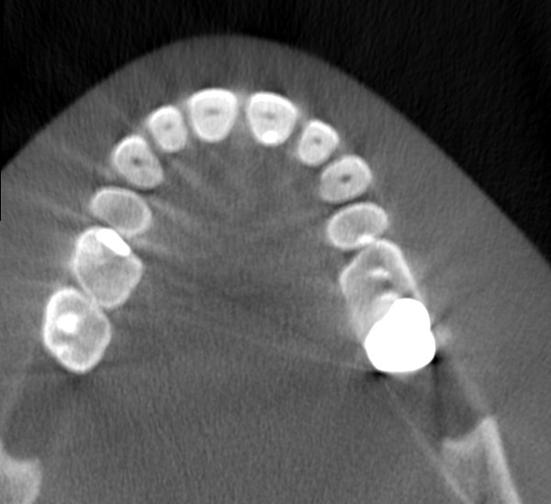}
    \caption{$Spike$}
    \label{fig:spike}
    \end{subfigure}
    \hfill
    \begin{subfigure}{0.32\textwidth}
    \includegraphics[width=\textwidth]{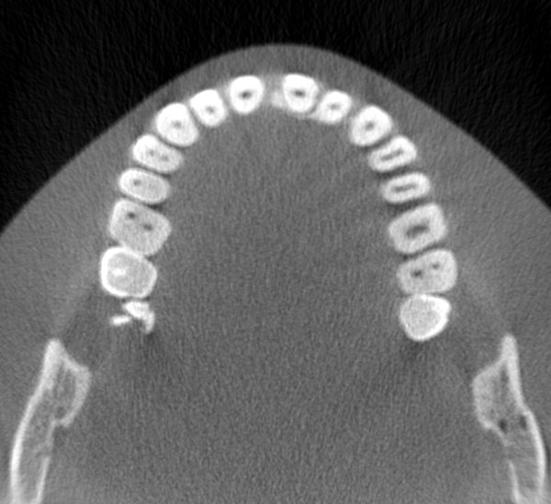}
    \caption{$Tom$}
    \label{fig:tom}
    \end{subfigure}
    \hfill
    \begin{subfigure}{0.32\textwidth}
    \includegraphics[width=\textwidth]{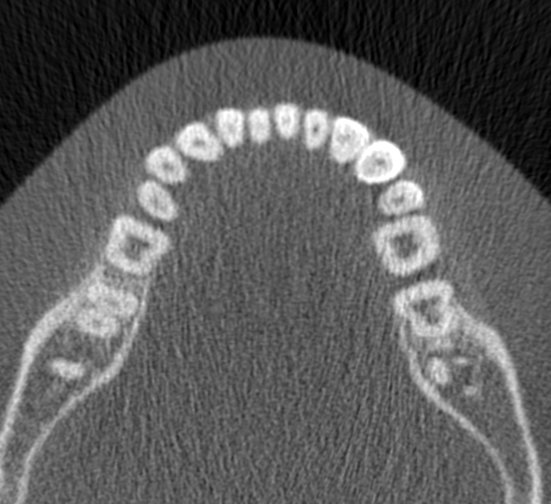}
    \caption{$Jerry$}
    \label{fig:jerry}
    \end{subfigure}
\caption{Slices showing the reconstruction of the phantoms}
\label{fig:phantoms}
\end{figure}

The imaging data (projection images) were acquired from the scans performed on a head CBCT Scanner at See Through s.r.l., Brusaporto, Italy. The distance between the x-ray source to the detector is $741 \ mm$. The detector pixel size is $0.24 \times 0.24 \ mm^2$ with a scan Field-of-View $105 \times 110 \ mm^2$. In this study, three head phantoms were used: $Jerry$, a  small phantom; $Tom$, slightly bigger compared to $Jerry$; and $Spike$, the biggest among the three. The $Spike$ phantom has metal objects like crowns and fillings which can create other non-motion-related artifacts e.g., beam hardening-induced over or under-estimation of attenuation and photon starvation-induced noise (see Figure \ref{fig:spike}). The phantoms are real skulls from deceased patients covered in a plastic resin that imitates the characteristics of the soft tissues in terms of X-ray attenuation. The use of phantoms is particularly well suited to motion detection and compensation in a setup similar to real-world clinical scenarios since CBCT visualizes details in the hard tissue (bones) rather than soft tissues. All the scans were performed for a full rotation ($360^{\circ}$) with a duration of 24 seconds.

\subsection{Motion Simulation} \label{ms}
\begin{figure}[htbp]
\centering
\begin{subfigure}{0.35\textwidth}
\includegraphics[width=\textwidth]{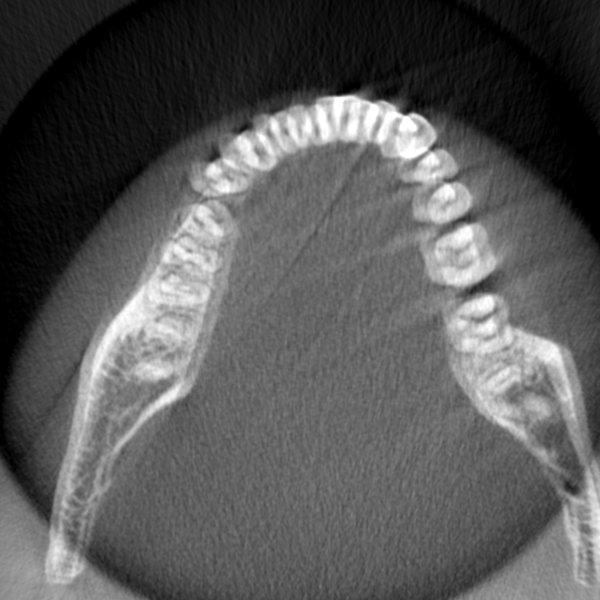}
\caption{Real Motion}
\label{fig:real-motion}
\end{subfigure}
\hspace{15mm}
\begin{subfigure}{0.35\textwidth}
\includegraphics[width=\textwidth]{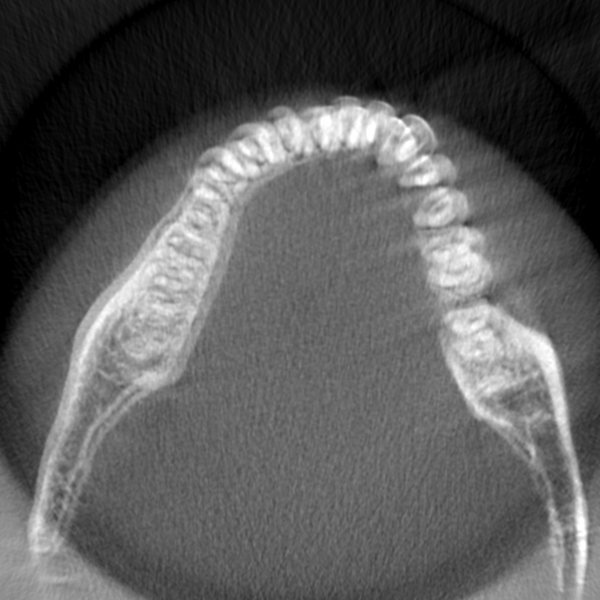}
\caption{Simulated Motion}
\label{fig:simul-motion}
\end{subfigure}
\caption{Comparison between artifacts from real and simulated motion for lateral rotation.}
\label{fig:tom-comparison}
\end{figure}

Motion-affected data can be acquired in different ways. Some researchers used robot programming and external movement devices to make the phantom move realistically while scanning \cite{spin2018ex}. Other research groups modeled the motion of the phantom with a digital simulation \cite{kim2016correction} which we adopted in this study. CBCT scanners typically describe the geometry of the acquisition by means of a set of projection matrices that encode the relative positions of the x-ray source and detector for each of the acquired projections. These matrices are necessary for reconstructing the volume of the object being scanned using either the Filtered Back Projection (FBP) technique or iterative techniques. These projection matrices can also be conveniently used to simulate motion by altering the parameters of the motion with rotation and translation. This approach allows the simulation of a broader set of motion patterns than in a few restricted clinical scenarios. Also, clinical data would need to be accompanied by ground-truth motion measurement, which would require a complex setting, difficult to implement in a clinical environment. This would likely prevent the collection of large amounts of data. A comparison between artifacts in real motion-affected data (see Section \ref{rm}) and data with digitally simulated motion is shown in Figure \ref{fig:tom-comparison}.

To simulate realistic motion, we rely on the motion studies of Spin-Neto \cite{spin2013cone,spin2016movement,spin2018ex,spin2015factors}. A variety of rigid motions can occur during CBCT scanning. Hence, we simulated four different motion types: nodding (\textbf{Nod}), tilting (\textbf{Tilt}), lateral rotation (\textbf{LR}), and tremor (\textbf{Trem}). The movement of the jaw is not considered as it is usually prevented by a fixed-bite block and chin rest which are standard accessories used in CBCT scanners. Additionally, three different motion patterns were considered; 1. the patient moves and stays in the final position for the rest of the scan (\textbf{sf}), 2. the patient moves and returns back to the initial position after some time (\textbf{ri}), 3. the patient moves and returns back to a particular state other than the initial position (\textbf{rd}) (see Table \ref{table:performance-analysis}). Spin-Neto et al. \cite{spin2016movement,spin2018ex} demonstrated that larger motions affecting image quality are confined to a smaller region. Therefore, the motion is estimated to be localized roughly in a range of 1-5 seconds for a full scan of 20 seconds. In this study, the motion is simulated in different parts of the scanned projections with varying amplitude between $3 \ - \ 8 \ degree/mm$.

\subsection{Data Augmentation}
The issue of data imbalance is a common problem in machine learning-based medical imaging approaches. In the case of motion detection, there is often a lack of positive data (with motion artifacts) and comparatively more negative data (without motion artifacts) \cite{sun2020motion,welch2020automatic}. In contrast, we have more positive data as the motion is artificially simulated but the negative data is reconstructed from only three head phantoms. To address this imbalance issue, we used data augmentation to increase the diversity of the negative data set. In most people, the dental arch (mandible and maxilla) is similar in shape with a slight variation in the size \cite{braun1998form,kairalla2014determining}. Thus we can generate more virtual phantoms than the three available ones by scaling with a factor between 0.8 and 1.2 to each of the three phantoms, and from these virtual augmented phantoms, we can generate more negative cases of motion-free scans. Additionally, random affine transforms with rotations up to  $5^{\circ}$ and translations up to $5 \ mm$ along X, Y, and Z axes are also applied. This helps to create a more balanced dataset in order to improve the performance of the model.

\subsection{Motion Artifacts Detection}
\begin{figure}[htbp]
  \includegraphics[width=\linewidth]{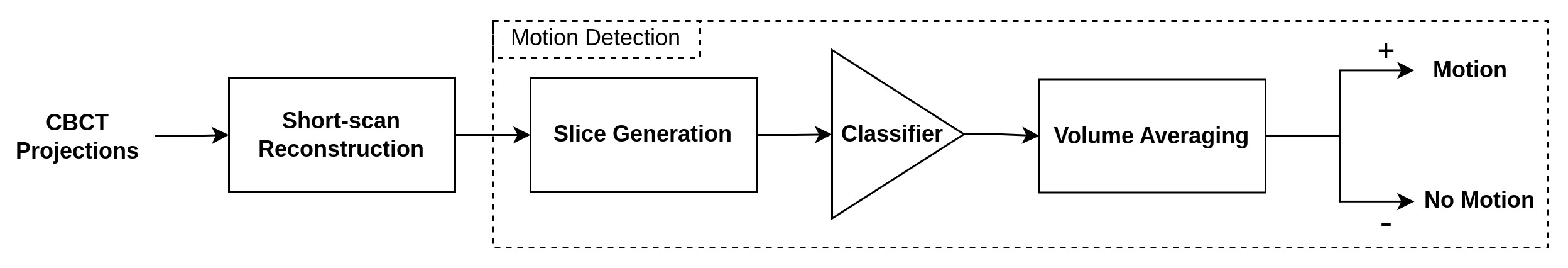}
  \caption{Framework of motion detection in CBCT volumes}
  \label{fig_pipeline}
\end{figure}

A deep learning-based approach is used in this study to detect motion artifacts in CBCT short-scan volumes. The framework has been implemented in three steps - Slice generation, Network classification, and Volume averaging as shown in Figure \ref{fig_pipeline}.

\subsubsection{Slice Generation}
The variability of features in the human skull bones is limited and statistically described well in the literature \cite{braun1998form,kairalla2014determining}. Basically, human skull bones and teeth are locally very similar, although they result in faces that look totally different to human observers. For instance, the sizes and shapes of teeth vary by very small amounts \cite{leung2018comparative}. Therefore teeth and jaw bones can be good features to train a generalized motion detection framework. We train the network to see a limited space of anatomical features, and the motion-affected images have distinctive features, different from motion-free anatomies. We apply this principle in a simplified setting since the teeth and jaw bones can be well seen in 2D axial slices extracted from the volume. Artifacts are often visible in most axial slices, making them a reliable choice for classification.

In this study, each short-scan volume is represented by 300 axial-view slices for classification. Using 3D data in the classification process can make the network architecture more complex and computationally expensive. A simpler network architecture, however, is often sufficient for classifying 2D data. 

\subsubsection{Network Classification}
Different CNN architectures have been used in this framework to analyze the performance of each model on our data. The models are constructed based on transfer learning using pre-trained ImageNet weights. In the training phase, Binary Cross Entropy Loss is used. All the networks are trained with Stochastic Gradient Descent (SGD) optimizer with a learning rate of 1e-3 and momentum of 0.9.  The slices are normalized and resized with respect to the input requirements of the network architectures.

\subsubsection{Volume Averaging}
In some slices where the direction of motion is parallel to the direction of strong edges, the artifacts due to motion may be less noticeable or compromised. This is because the strong edges can provide some structural information that can help to compensate for the motion-induced artifacts. To overcome this issue, volume averaging is performed. In the evaluation phase, the final prediction ($\textbf{y}_{final}$) is based on the mean outcome ($\textbf{y}_{pred}$) of the network per volume, i.e., the probability scores ($\textbf{y}$) for all the slices generated from a volume are averaged (Equation \ref{vol_avg}), and the average score is considered for the final prediction (Equation \ref{result}). The advantage of volume averaging is that, e.g., even if only 65\% of the slices are correctly predicted with an average probability of 0.8, the volume will still be classified correctly. 

\begin{equation}
 \textbf{y}_{pred} = \frac{\sum_{i} \textbf{y}_{i}}{N} 
 \label{vol_avg}
\end{equation}

\begin{equation}
 \textbf{y}_{final} = 
 \begin{cases}
 No \ Motion \ (-), &  \mbox{if } \textbf{y}_{pred}<0.5 \\ Motion \ (+), & \mbox{if } \textbf{y}_{pred}>=0.5 \\
 \end{cases} 
 \label{result}
\end{equation}

\section{Evaluation}

\subsection{Dataset}
The dataset consists of short-scan volumes reconstructed with the set of projection images that correspond to a $194^{\circ}$ angle. These short-scan volumes are derived from the full $360^{\circ}$ scan of the head phantoms (also including the augmented ones). We considered 4 default short-scan views to cover the full $360^{\circ}$ view. Since we applied a short-duration motion of 1-5 seconds for a scan of 24 seconds, there should always be at least one short-scan that is free from motion. The training dataset has 80 short-scan volumes, 40 volumes each from  $Tom$ and $Jerry$ phantoms. The volumes with and without motion artifacts are distributed equally in the training dataset. $Spike$ phantom is used for testing in order to study the robustness of our framework to detect motion in the presence of metal artifacts. The testing dataset consists of 200 short-scan volumes reconstructed from 50 projection data corresponding to 10 different motion types (Table \ref{table:performance-analysis}). 

\subsection{Experimental Results}
The experiments were conducted on a Windows-10 PC with Intel(R) Core(TM) i7-2.90 GHz processor, 16GB RAM, and NVIDIA RTX 3070 (8GB VRAM). The deep learning framework has been implemented using Pytorch with CUDA. The performance has been evaluated using the area under the Precision-Recall curve (AUC-PR) metric for both slices and volumes in the testing dataset for different network architectures. Five projection data are used per motion type and the short-scan reconstructions are labeled manually. Table \ref{table:performance-analysis} shows the average AUC-PR per motion type and overall average. ResNet performed well with respect to the number of slices being correctly predicted. However, the goal is to classify volumes, and in this case, EfficientNet V2 outperforms all other architectures. The results show that the framework is capable of specifically detecting motion artifacts in the presence of metal artifacts ($Spike$ phantom has metal crowns and fillings while $Tom$ \& $Jerry$ phantoms used in training don't have any metal objects or implants).

\begin{table}[H]
\centering
\caption{AUC-PR for the testing ($Spike$) dataset on different network architectures. The best figures for each motion type are in bold.}
\begin{adjustbox}{width=\textwidth}
\begin{tabular}{l|c|c|c|c|c|c|c|c|c|c}
\hline
\textbf{Motion} & \multicolumn{2}{c|}{\textbf{VGG}} & \multicolumn{2}{c|}{\textbf{DenseNet}} & \multicolumn{2}{c|}{\textbf{ResNet}} & \multicolumn{2}{c|}{\textbf{EfficientNet}} & \multicolumn{2}{c}{\textbf{EfficientNet V2}} \\ \cline{2-11}
\textbf{Tpye} & \textbf{Slices} & \textbf{Volume} & \textbf{Slices} & \textbf{Volume} & \textbf{Slices} & \textbf{Volume} & \textbf{Slices} & \textbf{Volume} & \textbf{Slices} & \textbf{Volume} \\ 
\hline
Nod-sf & \textbf{0.907} & \textbf{0.983} & 0.885 & 0.962 & 0.904 & \textbf{0.983} & 0.888 & 0.929 & 0.873 & 0.906 \\ 
Nod-ri & 0.829 & \textbf{1.000} & 0.801 & 0.857 & \textbf{0.844} & 0.950 & 0.791 & 0.917 & 0.786 & \textbf{1.000} \\ 
Nod-rd & 0.842 & 0.857 & 0.830 & 0.833 & \textbf{0.851} & 0.894 & 0.826 & 0.885 & 0.839 & \textbf{0.917} \\ 
Tilt-sf & 0.864 & 0.962 & 0.846 & 0.900 & \textbf{0.902} & \textbf{1.000} & 0.862 & 0.900 & 0.859 & \textbf{1.000} \\
Tilt-ri & 0.851 & 0.896 & 0.805 & 0.837 & \textbf{0.872} & \textbf{0.934} & 0.826 & 0.896 & 0.820 & 0.896 \\ 
Tilt-rd & 0.903 & \textbf{1.000} & 0.865 & 0.900 & \textbf{0.908} & 0.983 & 0.855 & 0.929 & 0.874 & \textbf{1.000} \\
LR-sf & 0.837 & 0.962 & 0.836 & 0.900 & \textbf{0.841} & 0.883 & 0.839 & 0.900 & 0.815 & \textbf{0.983} \\
LR-ri & 0.845 & \textbf{0.960} & 0.816 & 0.866 & \textbf{0.854} & 0.919 & 0.826 & 0.866 & 0.806 & 0.919 \\
LR-rd & 0.800 & 0.884 & 0.768 & 0.821 & \textbf{0.831} & \textbf{0.925} & 0.778 & 0.884 & 0.768 & \textbf{0.925} \\
Trem & 0.884 & \textbf{1.000} & 0.833 & 0.857 & \textbf{0.935} & \textbf{1.000} & 0.854 & 0.857 & 0.855 & \textbf{1.000} \\
\hline
\textbf{Average} & 0.856 & 0.950 & 0.828 & 0.873 & \textbf{0.874} & 0.947 & 0.834 & 0.896 & 0.830 & \textbf{0.955} \\ 
\hline
\end{tabular}
\end{adjustbox}
\label{table:performance-analysis}
\end{table}

\subsection{Validation on Real Data} \label{rm}
The framework has also been tested in real motion-affected data to evaluate its robustness in real-world scenarios. The data was acquired by moving the phantom during scanning to induce motion. A setup was used to move the phantom during the scan, thereby inducing a motion that is approximately equivalent to lateral rotation. Three scans were performed with motion at different time intervals and short-scan volumes were reconstructed for the 4 selected views. These volumes were then classified using different architectures and the results are presented in Table \ref{table:real-motion-results}. EfficientNet, EfficientNet V2, and VGG were able to cope-up with differences between real and simulated motions better than ResNet and DenseNet.

Simulated motion data is different from real-world scans in two ways. Firstly, the reconstruction is performed from simulated x-ray projections in which the noise due to scattering, photon starvation, and beam hardening are significantly less compared to acquired projections. Secondly, motion is modeled and not acquired from a real-world setup. Overall, the results show that the network is capable of inferring motion artifacts from real-world scans successfully, even though only simulated motion-affected data is used in training.

\begin{table}[H]
\centering
\caption{AUC-PR for real motion-affected data on different network architectures. The best figures are in bold.}
\label{table:real-motion-results}
\begin{adjustbox}{width=\textwidth}
\begin{tabular}{c|c|c|c|c|c|c|c|c|c|c}
\hline
\multirow{2}{*}{\textbf{Scans}} & \multicolumn{2}{c|}{\textbf{VGG}} & \multicolumn{2}{c|}{\textbf{DenseNet}} & \multicolumn{2}{c|}{\textbf{ResNet}} & \multicolumn{2}{c|}{\textbf{EfficientNet}} & \multicolumn{2}{c}{\textbf{EfficientNet V2}} \\ \cline{2-11}
~ & \textbf{Slices} & \textbf{Volume} & \textbf{Slices} & \textbf{Volume} & \textbf{Slices} & \textbf{Volume} & \textbf{Slices} & \textbf{Volume} & \textbf{Slices} & \textbf{Volume} \\ 
\hline
\textbf{Scan-1} & 0.934 & \textbf{1.000} & 0.920 & 0.875 & 0.918 & 0.875 & 0.937 & \textbf{1.000} & \textbf{0.970} & \textbf{1.000} \\
\textbf{Scan-2} & 0.931 & \textbf{1.000} & 0.894 & 0.875 & 0.907 & 0.875 & 0.933 & \textbf{1.000} & \textbf{0.941} & \textbf{1.000} \\
\textbf{Scan-3} & 0.855 & \textbf{1.000} & 0.801 & 0.750 & 0.802 & 0.750 & 0.855 & \textbf{1.000} & \textbf{0.896} & \textbf{1.000} \\
\hline
\textbf{Average} & 0.906 & \textbf{1.000} & 0.872 & 0.833 & 0.876 & 0.833 & 0.908 & \textbf{1.000} & \textbf{0.936} & \textbf{1.000} \\
\hline
\end{tabular}
\end{adjustbox}
\end{table} 

\section{Discussion}
The results show that the framework is able to reliably extract a short-scan CBCT reconstruction with sufficient diagnostic quality. This can be valuable in practice since no further correction steps are needed. Moreover, our framework is able to discriminate between motion artifacts and other artifact types (see Figure \ref{fig:spike}). This is a promising result since, in some cases, where too many projections are affected by motion, our framework can be selectively trained and possibly employed to recognize motion artifacts on a smaller subset of projections to provide a valid reference for correction algorithms. 

Furthermore, our approach shows that it is feasible to robustly train a network with simulated motion data and to achieve good performance on real motion-affected data. To our knowledge, there is no publicly available real dataset with measured motions for dental CBCT scanning.  Therefore, creating a dataset using simulation would be a valuable resource for the research community.



\section{Conclusion}
This paper presents a deep learning framework that focuses on detecting motion artifacts in short-scan dental CBCT reconstruction. Realistic motion simulation and augmentation are employed to address data scarcity. The framework with EfficientNet V2 as the backbone achieved the highest AUC-PR for the simulated motion data. Validation on real motion-affected data shows the robustness of the framework to detect motion in real-world scans. Further validation with patient data, would allow the performance to be evaluated in diverse populations, which is essential for developing a reliable and robust clinical tool.

\subsubsection{Acknowledgements}
The authors thank Michele Antonelli, Ivan Tomba, Andrea Delmiglio, Lorenzo Arici from See Through s.r.l., and Prof. Giovanni Di Domenico from the University of Ferrara, Italy for providing the projection data and the reconstruction libraries. The authors also wish to thank Prof. David Fofi from the University of Burgundy, France for his suggestions and support.

\bibliographystyle{splncs04}
\bibliography{mybibliography}

\appendix
\section{Short-scan Reconstructions for Simulated motion data}

\subsection*{\#\# Green indicates \textcolor{green}{no motion}; Red indicates \textcolor{red}{motion} artifacts}

\begin{figure}[H]
    \begin{subfigure}{0.23\textwidth}
    \includegraphics[width=\textwidth]{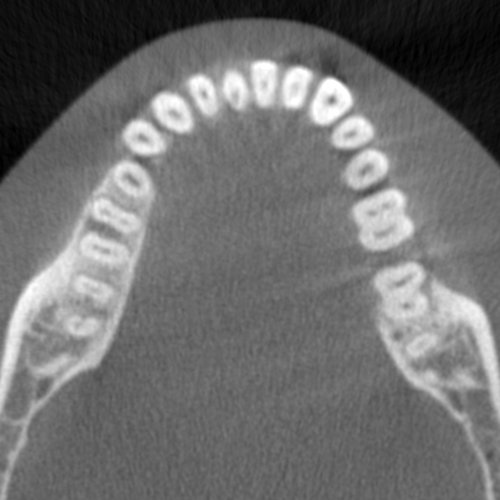}
    \caption{\textcolor{green}{View-1}}
    \label{fig:tom-1}
    \end{subfigure}
    \hfill
    \begin{subfigure}{0.23\textwidth}
    \includegraphics[width=\textwidth]{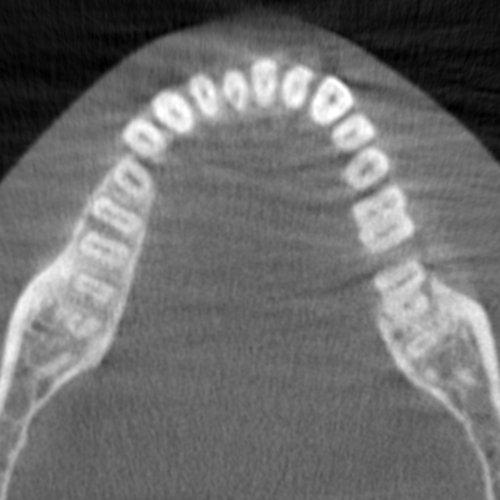}
    \caption{\textcolor{red}{View-2}}
    \label{fig:tom-2}
    \end{subfigure}
    \hfill
    \begin{subfigure}{0.23\textwidth}
    \includegraphics[width=\textwidth]{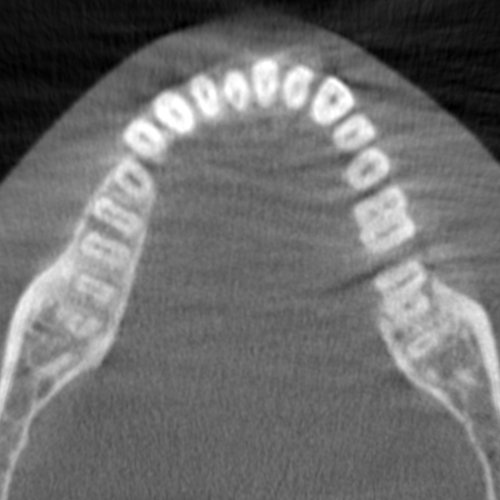}
    \caption{\textcolor{red}{View-3}}
    \label{fig:tom-3}
    \end{subfigure}
    \hfill
    \begin{subfigure}{0.23\textwidth}
    \includegraphics[width=\textwidth]{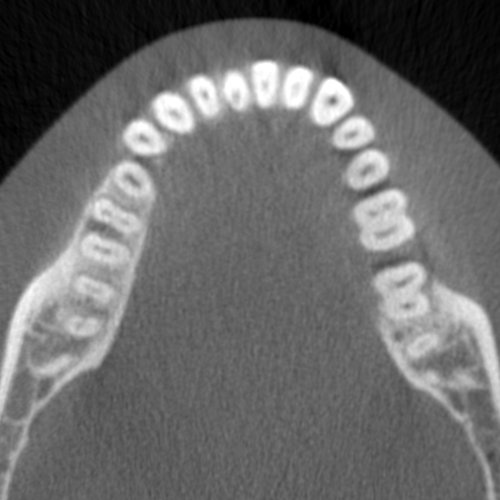}
    \caption{\textcolor{green}{View-4}}
    \label{fig:tom-4}
    \end{subfigure}
\caption{Short-scan regions for $Tom$ data with Tremble. }
\label{ffig:tom-trem}
\end{figure}

\begin{figure}[H]
    \begin{subfigure}{0.23\textwidth}
    \includegraphics[width=\textwidth]{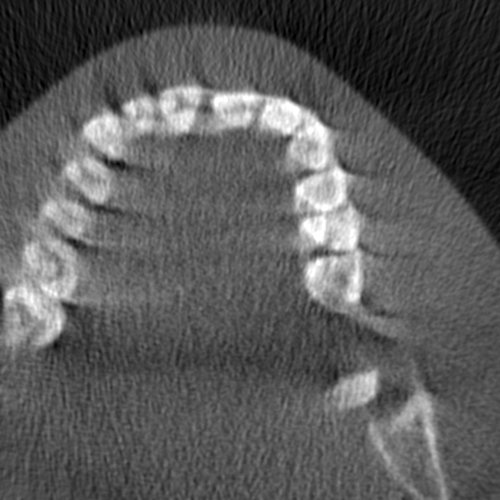}
    \caption{\textcolor{red}{View-1}}
    \label{fig:jerry-1}
    \end{subfigure}
    \hfill
    \begin{subfigure}{0.23\textwidth}
    \includegraphics[width=\textwidth]{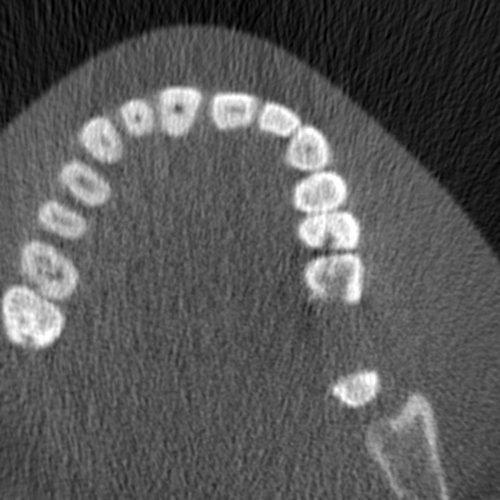}
    \caption{\textcolor{green}{View-2}}
    \label{fig:jerry-2}
    \end{subfigure}
    \hfill
    \begin{subfigure}{0.23\textwidth}
    \includegraphics[width=\textwidth]{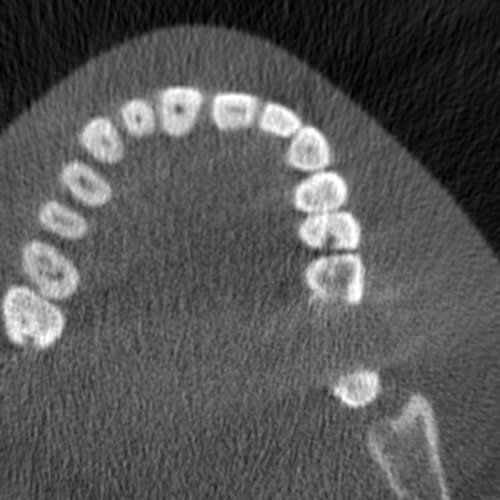}
    \caption{\textcolor{green}{View-3}}
    \label{fig:jerry-3}
    \end{subfigure}
    \hfill
    \begin{subfigure}{0.23\textwidth}
    \includegraphics[width=\textwidth]{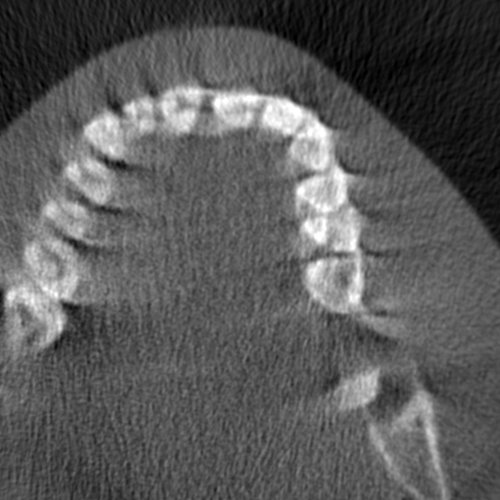}
    \caption{\textcolor{red}{View-4}}
    \label{fig:jerry-4}
    \end{subfigure}
\caption{Short-scan regions for $Jerry$ data with Tilting.}
\label{ffig:jerry-tilt}
\end{figure}

\begin{figure}[H]
    \begin{subfigure}{0.23\textwidth}
    \includegraphics[width=\textwidth]{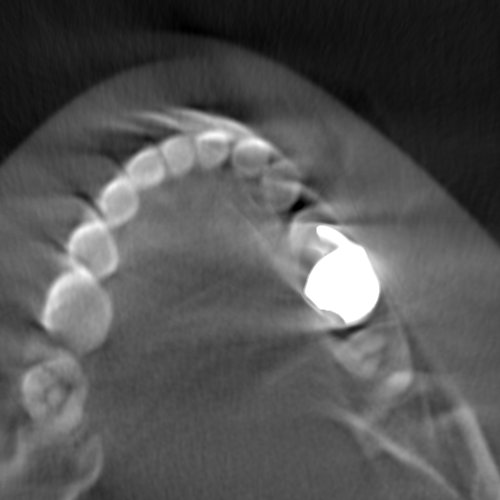}
    \caption{\textcolor{red}{View-1}}
    \label{fig:spike-1}
    \end{subfigure}
    \hfill
    \begin{subfigure}{0.23\textwidth}
    \includegraphics[width=\textwidth]{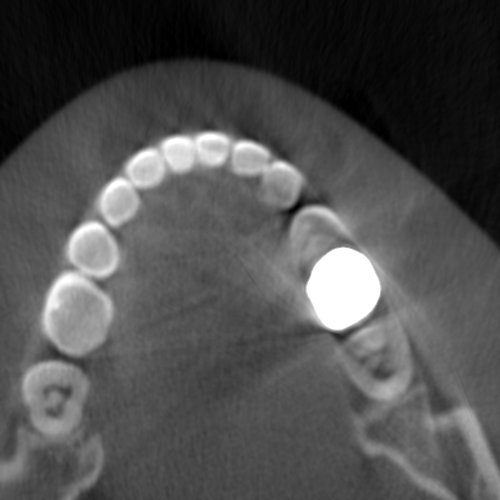}
    \caption{\textcolor{green}{View-2}}
    \label{fig:spike-2}
    \end{subfigure}
    \hfill
    \begin{subfigure}{0.23\textwidth}
    \includegraphics[width=\textwidth]{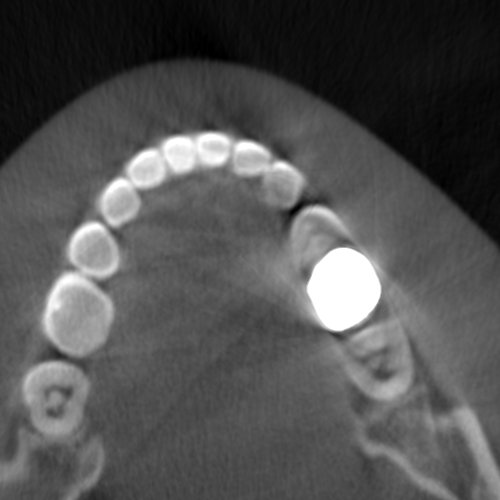}
    \caption{\textcolor{green}{View-3}}
    \label{fig:spike-3}
    \end{subfigure}
    \hfill
    \begin{subfigure}{0.23\textwidth}
    \includegraphics[width=\textwidth]{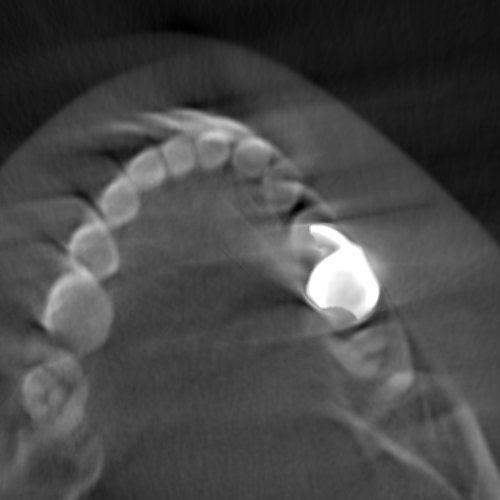}
    \caption{\textcolor{red}{View-4}}
    \label{fig:spike-4}
    \end{subfigure}
\caption{Short-scan regions for $Spike$ data with Nodding. Metal artifacts are visible in all the reconstruction views.}
\label{ffig:spike-nod}
\end{figure}

\section{Short-scan Reconstructions for Real motion-affected data}

\begin{figure}[H]
    \begin{subfigure}{0.23\textwidth}
    \includegraphics[width=\textwidth]{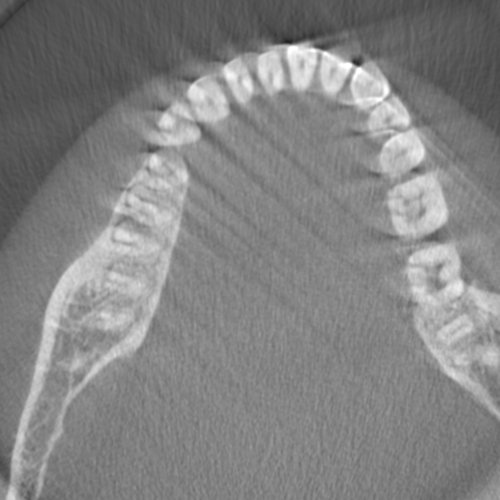}
    \caption{\textcolor{red}{View-1}}
    \label{fig:ssv1-1}
    \end{subfigure}
    \hfill
    \begin{subfigure}{0.23\textwidth}
    \includegraphics[width=\textwidth]{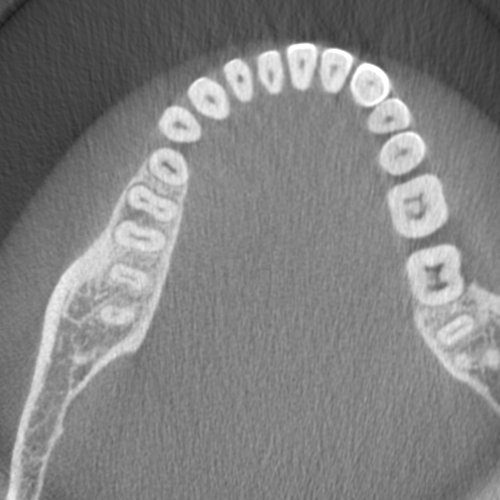}
    \caption{\textcolor{green}{View-2}}
    \label{fig:ssv1-2}
    \end{subfigure}
    \hfill
    \begin{subfigure}{0.23\textwidth}
    \includegraphics[width=\textwidth]{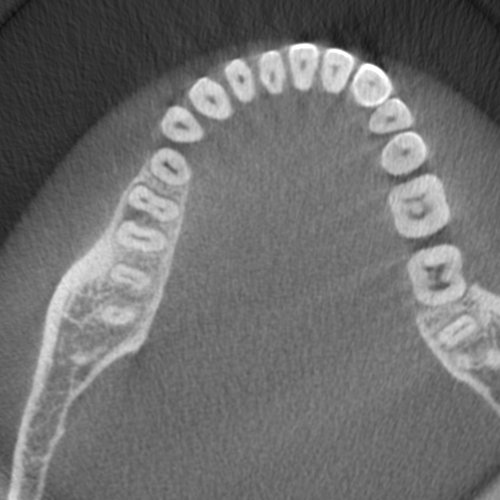}
    \caption{\textcolor{green}{View-3}}
    \label{fig:ssv1-3}
    \end{subfigure}
    \hfill
    \begin{subfigure}{0.23\textwidth}
    \includegraphics[width=\textwidth]{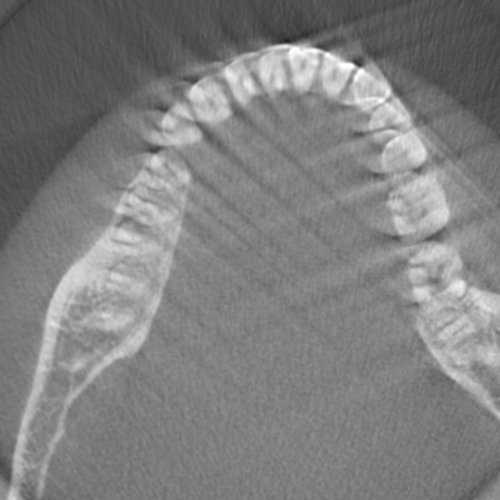}
    \caption{\textcolor{red}{View-4}}
    \label{fig:ssv1-4}
    \end{subfigure}
\caption{Short-scan regions for Scan 1.}
\label{ffig:real-motion-1}
\end{figure}

\begin{figure}[H]
    \begin{subfigure}{0.23\textwidth}
    \includegraphics[width=\textwidth]{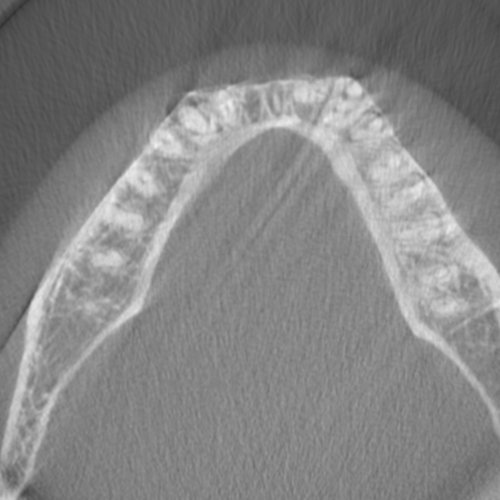}
    \caption{\textcolor{red}{View-1}}
    \label{fig:ssv2-1}
    \end{subfigure}
    \hfill
    \begin{subfigure}{0.23\textwidth}
    \includegraphics[width=\textwidth]{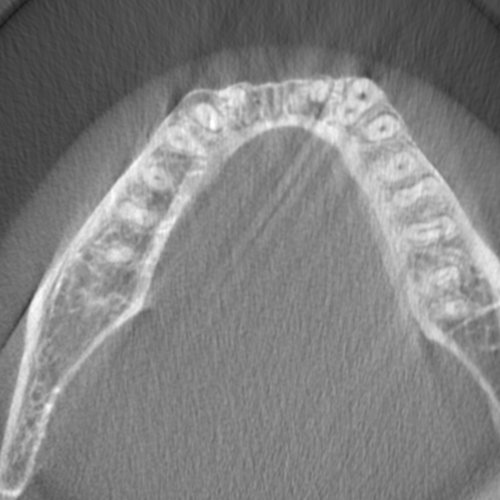}
    \caption{\textcolor{red}{View-2}}
    \label{fig:ssv2-2}
    \end{subfigure}
    \hfill
    \begin{subfigure}{0.23\textwidth}
    \includegraphics[width=\textwidth]{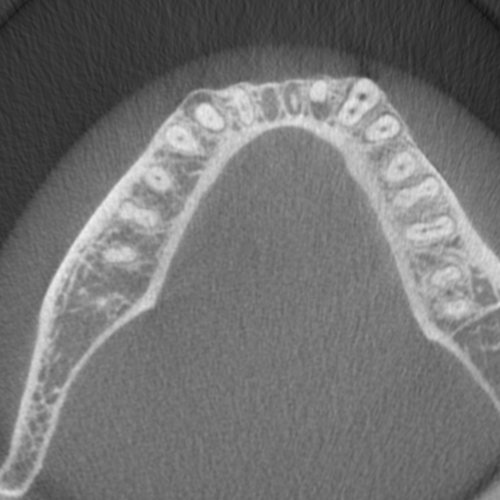}
    \caption{\textcolor{green}{View-3}}
    \label{fig:ssv2-3}
    \end{subfigure}
    \hfill
    \begin{subfigure}{0.23\textwidth}
    \includegraphics[width=\textwidth]{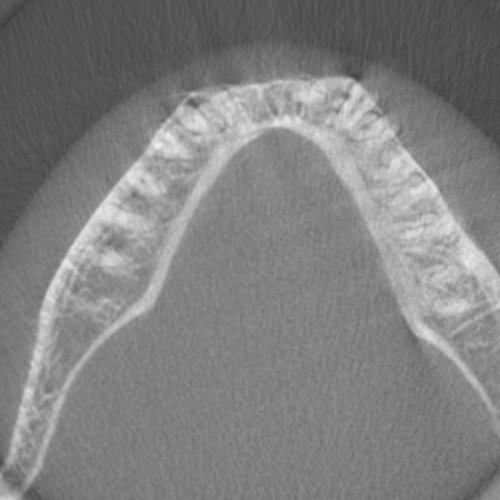}
    \caption{\textcolor{red}{View-4}}
    \label{fig:ssv2-4}
    \end{subfigure}
\caption{Short-scan regions for Scan 2.}
\label{ffig:real-motion-2}
\end{figure}

\begin{figure}[H]
    \begin{subfigure}{0.23\textwidth}
    \includegraphics[width=\textwidth]{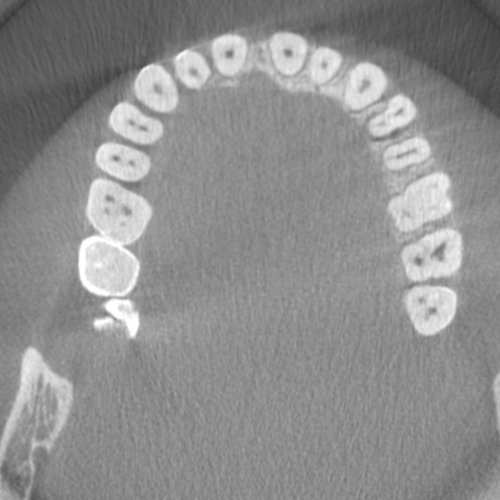}
    \caption{\textcolor{green}{View-1}}
    \label{fig:ssv3-1}
    \end{subfigure}
    \hfill
    \begin{subfigure}{0.23\textwidth}
    \includegraphics[width=\textwidth]{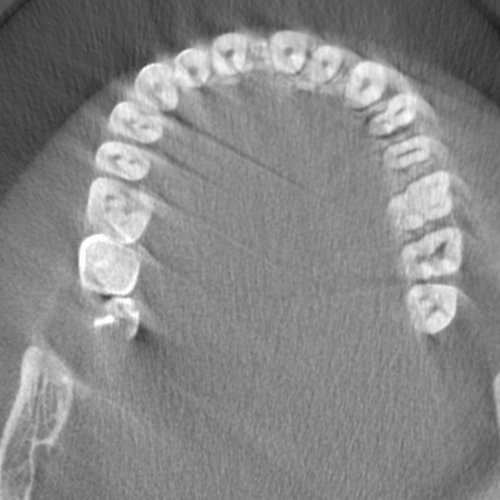}
    \caption{\textcolor{red}{View-2}}
    \label{fig:ssv3-2}
    \end{subfigure}
    \hfill
    \begin{subfigure}{0.23\textwidth}
    \includegraphics[width=\textwidth]{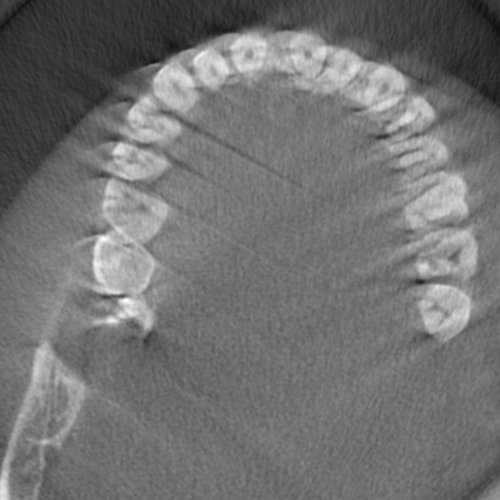}
    \caption{\textcolor{red}{View-3}}
    \label{fig:ssv3-3}
    \end{subfigure}
    \hfill
    \begin{subfigure}{0.23\textwidth}
    \includegraphics[width=\textwidth]{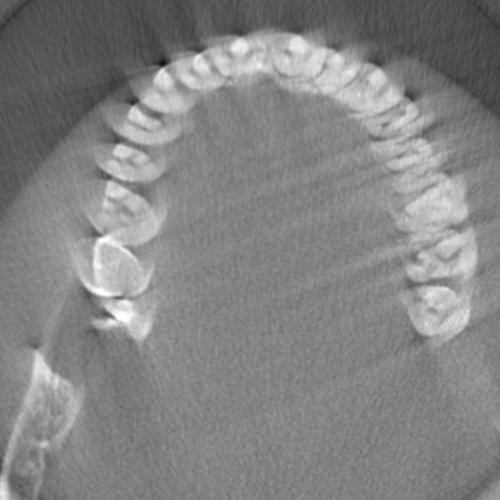}
    \caption{\textcolor{red}{View-4}}
    \label{fig:ssv3-4}
    \end{subfigure}
\caption{Short-scan regions for Scan 3.}
\label{ffig:real-motion-3}
\end{figure}

\end{document}